\documentclass[runningheads]{llncs}
\usepackage{graphicx}
\usepackage{multirow}

\begin{document}

\title{Transfer Learning for Ultrasound Tongue Contour Extraction with Different Domains}
\titlerunning{Transfer Learning for Ultrasound Tongue Contour Extraction}
\author{}
\author{M. Hamed Mozaffari \inst{*} \and Won-Sook Lee \inst{*}}
\authorrunning{M.H. Mozaffari et al.}
\institute{*School of Electrical Engineering and Computer Science, University of
Ottawa, 800 King-Edward Avenue, Ottawa, Ontario, Canada, K1N-6N5.}

\maketitle              % typeset the header of the contribution
\begin{abstract}
Medical ultrasound technology is widely used in routine clinical applications such as disease diagnosis and treatment as well as other applications like real-time monitoring of human tongue shapes and motions as visual feedback in second language training. Due to the low-contrast characteristic and noisy nature of ultrasound images, it might require expertise for non-expert users to recognize tongue gestures. Manual tongue segmentation is a cumbersome, subjective, and error-prone task. Furthermore, it is not a feasible solution for real-time applications. 
\\
In the last few years, deep learning methods have been used for delineating and tracking tongue dorsum. Deep convolutional neural networks (DCNNs), which have shown to be successful in medical image analysis tasks, are typically weak for the same task on different domains. In many cases, DCNNs trained on data acquired with one ultrasound device, do not perform well on data of varying ultrasound device or acquisition protocol. Domain adaptation is an alternative solution for this difficulty by transferring the weights from the model trained on a large annotated legacy dataset to a new model for adapting on another different dataset using fine-tuning. 
\\
In this study, after conducting extensive experiments, we addressed the problem of domain adaptation on small ultrasound datasets for tongue contour extraction. We trained a U-net network comprises of an encoder-decoder path from scratch, and then with several surrogate scenarios, some parts of the trained network were fine-tuned on another dataset as the domain-adapted networks. We repeat scenarios from target to source domains to find a balance point for knowledge transfer from source to target and vice versa. The performance of new fine-tuned networks was evaluated on the same task with images from different domains. 

\end{abstract}

\section{Introduction}

Ultrasound imaging is safe, relatively affordable, and capable of real-time performance. This technology has been utilized for many real-time medical applications. Recently, ultrasound is used for visualizing and characterizing human tongue shape and motion in a real-time speech to study healthy or impaired speech production in applications such as visual second language training \cite{gick2008ultrasound} or silent speech interfaces \cite{csapo2017dnn}. However, it requires expertise for non-expert users to recognize tongue shape and motion in noisy and low-contrast ultrasound data. To address this problem and to have a quantitative analysis, tongue surface (dorsum) can be extracted, tracked, and visualized superimposed on the whole tongue region. Delineating the tongue surface from each frame is a cumbersome, subjective, and error-prone task. Moreover, the rapidity and complexity of tongue gestures have made it a challenging task, and manual segmentation is not a feasible solution for real-time applications. 
\\
Over the years, several image processing techniques, such as active contour models \cite{li2005automatic}, have shown their capability for automatic tongue contour extraction. In many of those traditional methods, manual labelling, initialization, monitoring, and manipulation are frequently needed. Furthermore, those methods are computationally expensive whereas the image gradient should be calculated for each frame \cite{laporte2018multi}. In recent years, convolutional neural networks (CNN) have been the method of choice for medical image analysis with outstanding results \cite{litjens2017survey}. In a few studies, automatic tongue contour extraction \cite{zhu2018automatic} and tracking \cite{mozaffari2018guided} using CNN have been already investigated. In spite of their excellent results on a specific dataset from one ultrasound device, the generalization of those methods on test data with different distributions from different ultrasound device is often not investigated and evaluated. Although ultrasound tongue datasets have different distributions, there is always a correlation between movements of the tongue and its possible positions in the mouth. Therefore, domain adaptation might provide a universal solution for automatic, real-time tongue contour extraction, applicable to the majority of ultrasound datasets.
\\
In transfer learning \cite{pan2010survey}, a marginal probability distribution $P(X)$, where $X = \left \{ x_{1}, ..., x_{n} \right \}$ defined on a feature space of $R$ can be used for expressing a domain $D$. On a specific domain $D = \left \{R, P(X) \right \}$, comprises of pair of a label space $Y$ and an objective function $f(\cdot)$ in a form of $T = \left \{ Y, f(\cdot) \right \}$, the objective function $f(\cdot)$ can be optimized and learned the training data, which consists of pairs $\left \{x_{i}, y_{i} \right \}$, where $x_{i} \in  X$ and $y_{i} \in Y$ in a supervised fashion using one CNN model. After termination of the optimization process, the trained CNN model denoted by $\tilde{f}(\cdot)$ can predict the label of a new instance $x$. Transfer learning is defined as the procedure of enhancing the target prediction function $f_{T}(\cdot)$ in $D_{T}$ using the information in $D_{S}$ and $T_{S}$, whereas a source domain $D_{s}$ with a learning task $T_{S}$ and a target domain $D_{T}$ with learning task $T_{T}$ are given and $D_{S} \neq D_{T}$, or $T_{S} \neq T_{T}$ \cite{ghafoorian2017transfer}. Therefore, the prediction function $\tilde{f}_{ST}(\cdot)$ first is trained on the source domain $D_{S}$ and then fine-tuned for the target domain $D_{T}$. Conversely, $\tilde{f}_{TS}(\cdot)$ is initially trained for the target task, and then it is domain-adapted on the source dataset. 
\\
Fully convolutional networks (FCNs) consist of consecutive convolutional and pooling layers (encoder), and one up-sampling layer (decoder) was successfully exploited for the semantic segmentation problem in a study by \cite{long2015fully}. Due to the loss of information in polling layers, only one layer of up-sampling cannot retrieve the input-sized resolution in the output prediction map. Concatenation of feature maps from deconvolutional layers (DeconvNet) \cite{noh2015learning} and encoder layers in U-net \cite{ronneberger2015u} improved significantly the segmentation accuracy. The encoder of U-net learns simple visual image features especially in the first few layers, while the decoder aims to reconstruct the input-sized output prediction map from the complicated, abstract, and task-dependent features of the last layer of the encoder. Although encoder-decoder models like U-net have been used for tongue contour extraction, still it is not obvious how much knowledge is preserved during the transfer learning process for domain adaptation. 
\\
In this study, the performance of the U-net in different scenarios was analyzed to answer some fundamental questions in domain adaptation. We investigated how many layers from the decoder part should be fine-tuned to achieve the best segmentation accuracy in both the source and target domain at the same time (we called that a balanced point). Furthermore, the efficacy of dataset size in target domain along with the skip operation and concatenation on the performance of the U-net were explored on the problem of ultrasound tongue contour extraction.

\section{Materials and Method}
\subsection{Dataset}
Ultrasound video frames were randomly selected from recorded videos of a linear transducer connected to a Sonix Tablet ultrasound device at the University of Ottawa as well as videos from Seeing Speech project \cite{lawson2015seeing}. Using informed undersampling method \cite{berry2012training}, we generated two 2050 image datasets with different distributions, dataset I (uOttawa) and dataset II (SeeingSpeech). In this method, an average intensity image is calculated over the entire dataset, then one score is assigned for each frame depends on its intensity distance to the average image. After sorting the data by ranking order, we selected 2000 frames with the highest rank (high variance) and 50 images with the lowest grade (low variation). 
\\
Ground truth labels corresponding to each data was annotated semi-automatically by two experts using our custom annotation software. Off-line augmentation comprises of natural transformations in ultrasound data (e.g., horizontal flipping, restricted rotation and zooming) was employed to create larger datasets (50K for each). We split each dataset into training, validation, and test sets using $\%90/5/5$ ratios.

\subsection{Network Architecture and Training}
\begin{figure}[h!]
\centering
 \makebox[\textwidth]{\includegraphics[width=.6\paperwidth]{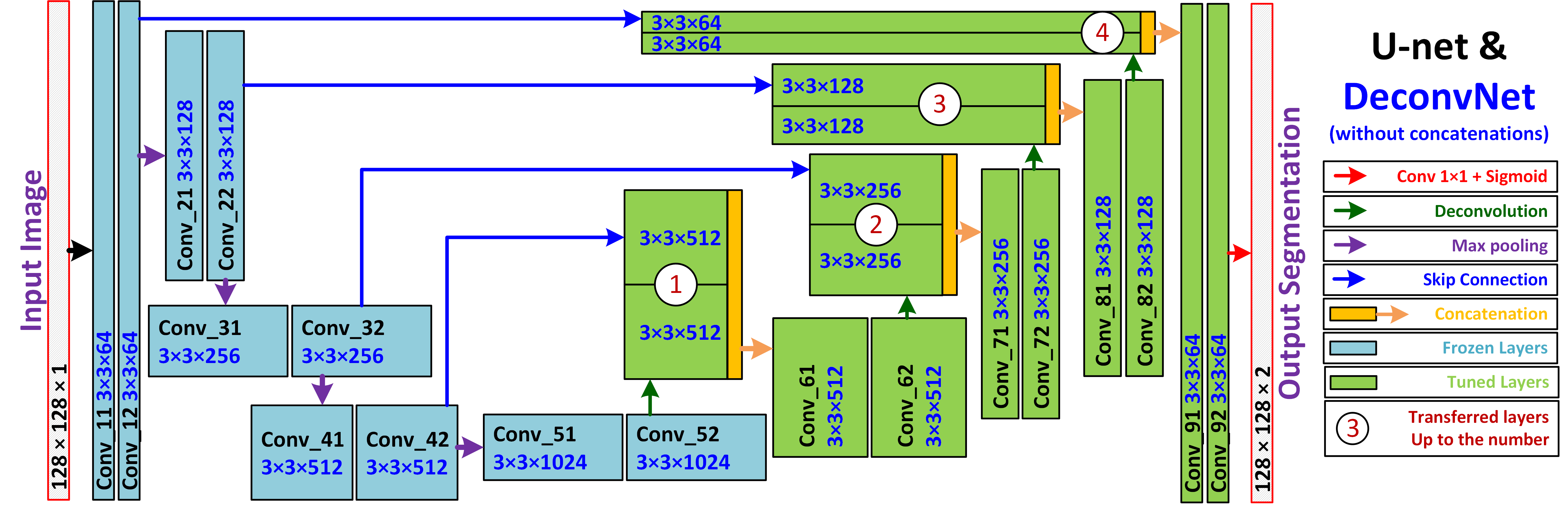}}
\caption{An overview of network structures. Numbers in circles show several scenarios for finding the best model for domain adaptation.}
\label{fig1}
\end{figure}

Fully convolutional networks (FCNs) can be considered as dense classification networks (e.g., VGG-nets) with consecutive convolutional and pooling layers such that a fully convolutional layer substitutes the fully connected layer (e.g., softmax in the last layer). Similarly, DeconvNet is an FCN network with several deconvolutional layers in the up-sampling path. In U-net \cite{ronneberger2015u} which is a DeconvNet architecture, feature maps (coarse contextual information) skips from each down-sampling layer to concatenate with deconvolutional layers for increasing the accuracy of output segmentation. Structural details of U-net have been presented in Fig. \ref{fig1}. 

The DeconvNet comprises of 9 double convolutional layers of $3\times3$ filters with Rectified Linear Unit (ReLU) activation function as non-linearity. Activations of all layers were normalized using batch normalization layers to speed up the convergence. In the downsampling path, there are four max-pooling layers for the sake of translation invariance and saving memory by decreasing learnable parameters. In contrast, in the up-sampling path, there are four deconvolutional layers which retrieve the original receptive filed and spatial resolution. Finally, the high-level reasoning is done by a fully convolutional layer at the end layer. 
\\
Network models were deployed using the publicly available TensorFlow framework on Keras API as the backend library. For initialization of network parameters, randomly distributed values have been selected. Adam optimization was chosen with a fixed momentum value of 0.9 for finding the optimum solution on a binary cross-entropy loss function. Each network model was trained using the mini-batch method employing one NVIDIA 1080 GPU unit which was installed on a Windows PC with Core i7, 4.2 GHz speed, and 32 GB of RAM. Our results from hyper-parameter tuning revealed that, besides network architecture size, learning rate has the most significant effect on the performance of each architecture in terms of accuracy. Testing fixed and scheduled decaying learning rate showed that the variable learning rate might provide better results, but it requires different initialization of decay factor and decay steps. Therefore, for the sake of a fair comparison, we only reported results using fixed learning rates. To alleviate the over-fitting problem, we regularized our networks by drop-out rate of 0.5. Networks were trained for a maximum of five epochs, each of which for 5000 iterations and mini-batch size of 10.

\subsection{Domain Adaptation Scenarios}
Models for $\tilde{f}_{ST}(.)$ were built from several scenarios with transferring the learned weights from $\tilde{f}_{S}$ when we froze the encoder and some parts of the decoder sections. Specifically, in scenario I, we transferred weights of the whole encoder as well as portions of the DeconvNet decoder as $\tilde{f}_{S}$ which was learned on the dataset I ($D_{S}$), then we froze those sections up to the $i$th deconvolutional layer and fine-tuned the remaining ($4-i$) deconvolutional layers using the Dataset II ($D_{T}$) (see the circled numbers in Fig \ref{fig1}). In scenario II, we investigated the opposite transferring case by switching source and target datasets to build model $\tilde{f}_{TS}(.)$ to see the effect of negative transferring. In similar scenarios, we repeated the same experiments by considering the impact of skip operator and concatenation in U-net to investigate the effect of transferring knowledge by injecting feature maps to the decoder from the encoder.

\section{Experimental Results}
To evaluate models, we investigated and compared different scenarios of tongue contour extraction as described in the previous section. In each situation, we first trained the whole DeconvNet and U-net on the source domains (named base models) and directly apply them on two source and target domains to see the weakness of each model in terms of generalization from one domain to another. From table \ref{tab1}, as it was anticipated, in both scenarios, base networks predicted better instances for their source domains than their target domains. The result of each scenario related to DeconvNet and U-net have been presented in table \ref{tab1}. 
\\
Results of the table reveal that on average fine-tuning the whole decoder section is the best for achieving the best accuracy in target domain while the negative transferring can be seen clearly in these cases. For instance, in the scenario I, in case of U-net base model, it achieved a Dice coefficient of 0.6884 for the source domain and 0.4664 for the target domain. At the same time, when the whole decoder fine-tuned a better Dice coefficient of 0.6306 was achieved in the target domain and 0.5818 in the source domain. As it can be seen, by freezing more layer in the decoder section (conv7, conv8, and conv9) the difference between the Dice coefficient values in source and target domains significantly increases. For the case of DeconvNet, this is not true and the difference decrease in higher layers. Table \ref{tab1} also indicate considerable result improvement in the scenario I for the U-net compare to the DeconvNet due to the concatenation and skip operation.

\begin{table}[h!]
\centering
\caption{Quantitative results of each scenario. Negative knowledge transferring can be seen in the two first columns for both models.}
\begin{tabular}{l|l|l|l|l|l|l|l|l|l|l|l}
\hline
\multicolumn{2}{l|}{\multirow{2}{*}{\begin{tabular}[c]{@{}l@{}}Scenario I\\ $D_{S} \rightarrow D_{T}$ \end{tabular}}}  & DecNet          & \multicolumn{4}{l|}{Transferred up to}              & U-net           & \multicolumn{4}{l}{Transferred up to}     \\ \cline{3-12} 
\multicolumn{2}{l|}{}                                                                          & Base            & Encod & conv 7          & conv 8 & conv 9 & Base            & Encod        & conv 7 & conv 8 & conv 9 \\ \hline
\multirow{2}{*}{\begin{tabular}[c]{@{}l@{}}Test \\ $D_{S}$\end{tabular}}           & Loss         & \textbf{0.2888} & 0.3212          & 0.3296          & 0.3711 & 0.4434 & \textbf{0.2269} & 0.3034          & 0.3203 & 0.3431 & 0.2464 \\ \cline{2-12} 
                                                                                & Dice         & \textbf{0.6584} & 0.5957          & 0.5744          & 0.5768 & 0.5891 & \textbf{0.6884} & 0.5818          & 0.5777 & 0.6274 & 0.6213 \\ \hline
\multirow{2}{*}{\begin{tabular}[c]{@{}l@{}}Test \\ $D_{T}$\end{tabular}}          & Loss         & 0.4999          & 0.3573          & \textbf{0.3252} & 0.4100 & 0.4513 & 0.4805          & \textbf{0.3129} & 0.3600 & 0.3963 & 0.4627 \\ \cline{2-12} 
                                                                                & Dice         & 0.5011          & \textbf{0.5779} & 0.5760          & 0.5131 & 0.5622 & 0.4664          & \textbf{0.6306} & 0.5074 & 0.5558 & 0.3808 \\ \hline
\multicolumn{2}{l|}{\multirow{2}{*}{\begin{tabular}[c]{@{}l@{}}Scenario II\\ $D_{T} \rightarrow D_{S}$\end{tabular}}} & DecNet          & \multicolumn{4}{l|}{Transferred up to}              & U-net           & \multicolumn{4}{l}{Transferred up to}     \\ \cline{3-12} 
\multicolumn{2}{l|}{}                                                                          & Base            & Encod & conv 7          & conv 8 & conv 9 & Base            & Encod        & conv 7 & conv 8 & conv 9 \\ \hline
\multirow{2}{*}{\begin{tabular}[c]{@{}l@{}}Test\\ $D_{T}$\end{tabular}}           & Loss         & \textbf{0.3286} & 0.4423          & 0.4981          & 0.4856 & 0.4332 & \textbf{0.3736} & 0.5100          & 0.5777 & 0.5591 & 0.6015 \\ \cline{2-12} 
                                                                                & Dice         & \textbf{0.6570} & 0.4685          & 0.3977          & 0.3611 & 0.4732 & \textbf{0.5901} & 0.4052          & 0.2931 & 0.3032 & 0.2301 \\ \hline
\multirow{2}{*}{\begin{tabular}[c]{@{}l@{}}Test \\ $D_{S}$\end{tabular}}           & Loss         & 0.4831          & \textbf{0.2571} & 0.2997          & 0.2986 & 0.2908 & 0.4253          & \textbf{0.2635} & 0.3363 & 0.3296 & 0.3270 \\ \cline{2-12} 
                                                                                & Dice         & 0.5299          & \textbf{0.6378} & 0.5816          & 0.5659 & 0.5921 & 0.5283          & \textbf{0.6211} & 0.5263 & 0.5361 & 0.5268 \\ \hline
\end{tabular}
\label{tab1}
\end{table}

\begin{figure}[h!]
\centering
 \makebox[\textwidth]{\includegraphics[width=\textwidth]{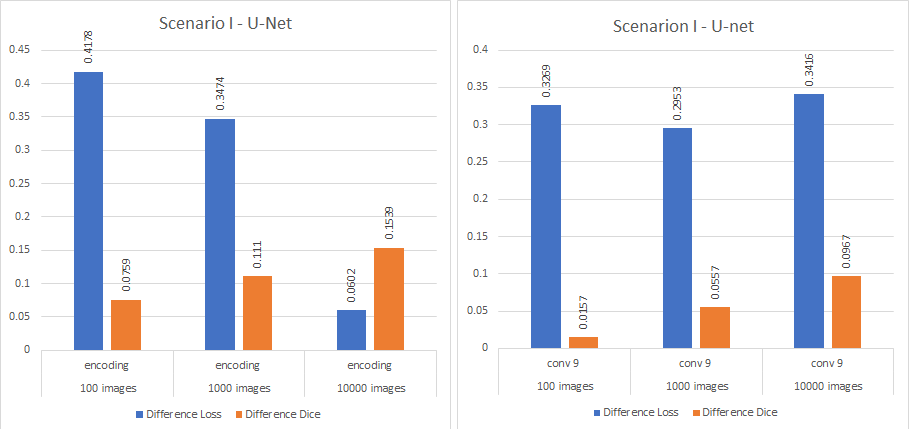}}
\caption{Effect of increasing dataset sizes on the accuracy of two transferred models in scenario I. Each column shows the difference of cross-entropy loss and dice-coefficient between source and target domains.}
\label{fig2}
\end{figure}

To identify the sufficient size of the target dataset for transfer learning, in separate experiments, we turned two transferred U-net models (encoder and conv 9) on three datasets with sizes of 100, 1000, and 10000. We used the same network architecture and training procedure among the different experiments. Figure \ref{fig2} shows the difference values between dice coefficients and cross-entropy losses in source and target domains for scenario I. It can be seen that more data samples enhance the performance of each model in terms of accuracy.

\begin{figure}[ht!]
\centering
\includegraphics[width=\textwidth]{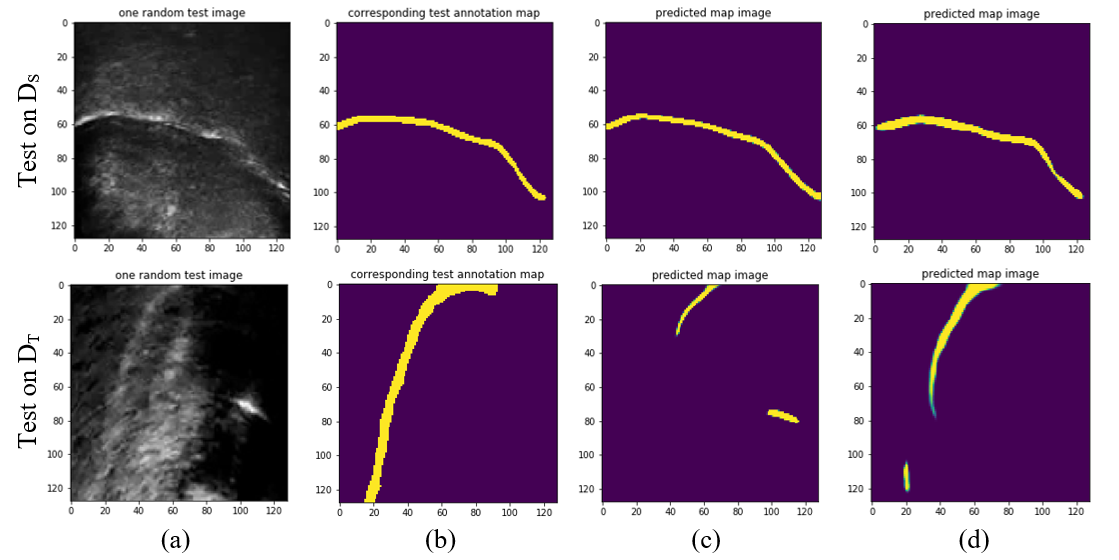}
\caption{Prediction results of U-net in scenario I $D_{S} \rightarrow D_{T}$. (a) sample data, (b) corresponding ground truth labels, (c) prediction result of $\tilde{f}_{S}$ (U-net base), (d) prediction result of $\tilde{f}_{ST}$ when the whole decoder was fine-tuned on $D_{T}$.}
\label{fig3}
\end{figure}

Fig \ref{fig3} illustrates the qualitative results of the scenario I for U-net model applied on a test instance. The U-net (base model) was trained on the set of images from the source domain ($\tilde{f}_{S}$), achieved a Dice coefficient of 0.65 and Binary cross-entropy loss of 0.28 while for the same model, the value of Dice score and loss for the target domain was 0.50 and 0.49 without fine-tuning. It means that although the result of the target domain is not significant, U-net base model can still predict instances in both source and target domains. Nevertheless, in case of real-time testing when some frames contain rapid tongue movement along with noisy dorsum region with artifacts (see Fig. \ref{fig3}.c), the model fails in prediction for the target domain. 
\\
Our experimental results revealed that fine-tuning of the whole decoder of the U-net alleviates this problem significantly. For instance, dice score and loss values approximately become 0.58 and 0.34  for both source and target domains when the whole decoder fine-tuned on the target domain. In general, we observed a balance point for the number of refined layers considering both source and target domains. On the balance point, the model can achieve similar acceptable results in both source, and target domains whereas the segmentation accuracy is worse than the model's performance on only one domain (see Fig. \ref{fig3}.d).
\section{Discussion and Conclusions}
In transfer learning literature, researchers usually focus on finding $\tilde{f}_{S}$ which demonstrates a decent performance on a source domain $D_{S}$. Then they try domain adaptation from source to target $D_{T}$ task to find $\tilde{f}_{ST}$. However, a reliable and universal method is the one which can provide acceptable results in the opposite path from target to source domain as well. Our experimental results showed that there is a balance point for U-net model where it can provide reasonable predictions on both the source and the target domain ($\tilde{f}_{ST} \sim \tilde{f}_{TS}$). For instance, transferring the whole decoder of U-net on the target domain, it provided Binary cross-entropy loss values of 0.3034 and 0.3129 for source and target test data.
\\
Furthermore, qualitative study shows that domain adaptation can improve segmentation result for frames with significant noise and artifacts. Impact of using skip operator and concatenation and increasing dataset size in target domain indicate a slight improvement in final results. In contrast with other research fields with large datasets, the size of a usual ultrasound tongue dataset is not more than $\sim 200K$ frames, and it makes more sense to fine-tune one model on several datasets to find the knowledge balance point as a universal model for use in real-time applications on various ultrasound devices. Using smaller learning rates in the target domains might increase the accuracy of the source and target domain segmentation further on the balance point. 

\bibliographystyle{splncs04}

\end{document}